\documentclass{article}

\usepackage[preprint,nonatbib]{neurips_2024}

\usepackage{enumitem}
\usepackage[utf8]{inputenc} 
\usepackage[T1]{fontenc}    
\usepackage{url}            
\usepackage{booktabs}       
\usepackage{amsfonts}       
\usepackage{nicefrac}       
\usepackage{microtype}      
\usepackage{xcolor}         
\newcommand{\ours}{\texttt{NeuronGen}\xspace}

\usepackage{times}
\usepackage{wrapfig}
\usepackage{epsfig}
\usepackage{graphicx}
\usepackage{amsmath}
\usepackage{amssymb}
\usepackage{xspace}
\usepackage[linesnumbered,ruled,vlined]{algorithm2e}
\usepackage{algpseudocode}
\usepackage{tabularx}
\usepackage{multirow}
\usepackage{subcaption}
\usepackage{booktabs}
\usepackage{xspace}
\SetKwInput{Input}{Input}
\SetKwProg{kwServer}{ServerUpdate}{:}{}
\SetKwProg{kwClient}{ClientUpdate}{:}{}

\SetKwProg{kwServer}{\textcolor{blue}{Server Update}}{}{}
\SetKwProg{kwClient}{\textcolor{blue}{Client Update}}{}{}

\usepackage{makecell}
\usepackage{bbm}
\usepackage{wrapfig}
\usepackage{graphicx}
\usepackage{ifthen}
\usepackage{xspace}
\usepackage{upgreek}
\usepackage{pifont}
\usepackage{colortbl}
\usepackage{cite}
\usepackage{multirow}
\usepackage{dsfont}
\usepackage[numbers]{natbib} 

\title{\ours: Neural Network Parameter Generation via Large Language Models}

\author{%
  Jiaqi Wang$^{1*}$ \;\;\;  
  Yusen Zhang$^{2*}$ \;\;\;\;   
  Xi Li$^{3}$\thanks{The three authors contributed equally to this work.}\\
  $^1$Auburn University,
  $^2$The Pennsylvania State University, 
  $^3$University of Alabama at Birmingham\\
  \texttt{jqwang@auburn.edu},
  \texttt{yfz5488@psu.edu}, \texttt{xiliuab@uab.edu} 
}

\begin{document}

\maketitle

\begin{abstract}
Acquiring the parameters of neural networks (NNs) has been one of the most important problems in machine learning since the inception of NNs. Traditional approaches, such as backpropagation and forward-only optimization, acquire parameters via iterative data fitting to gradually optimize them. This paper aims to explore the feasibility of a new direction: acquiring NN parameters via large language model generation. We propose \texttt{NeuronGen}, a generalized and easy-to-implement two-stage approach for NN parameter generation conditioned on descriptions of the data, task, and network architecture. Stage one is Parameter Reference Knowledge Injection, where LLMs are pretrained on NN checkpoints to build foundational understanding of parameter space, whereas stage two is Context-Enhanced Instruction Tuning, enabling LLMs to adapt to specific tasks through enriched, task-aware prompts. Experimental results demonstrate that \texttt{NeuronGen} effectively generates usable NN parameters. Our findings highlight the feasibility of LLM-based NN parameter generation and suggest a promising new paradigm where LLMs and lightweight NNs can coexist synergistically\footnote{The codes will be public after being accepted}.
\end{abstract}

\section{Introduction}
The acquisition of parameters in neural networks is primarily driven by gradient-based optimization. A typical approach to obtain a functional neural network involves forward and backward propagation over training data using gradient descent~\cite{rumelhart1986learning}. This classical framework has demonstrated remarkable effectiveness across various neural network architectures and domains, including but not limited to computer vision~\cite{krizhevsky2017imagenet,he2016deep}, time-series analysis~\cite{graves2012long}, and natural language processing~\cite{vaswani2017attention,devlin2019bert}.

Recently, several studies utilize diffusion models to obtain neural network parameters~\cite{wang2024neural,jin2024conditional,peebles2022learning}. Limited by the diffusion technique itself, most of the approachaes suffer from slow sampling speeds~\cite{ho2020denoising} and limited controllability~\cite{dhariwal2021diffusion}. Also, many of them generate parameter distributions rather than directly usable model parameters.

In this paper, \textbf{we explore the possibility and feasibility of a novel neural network parameter acquisition approach through LLM-based generation.}. Many research work~\cite{carlini2019secret,shokri2017membership,song2019auditing} have shown that the paramaters may carry the information of the data and there could be a hidden mapping between the parameters and data. 
As we know, LLMs have demonstrated remarkable capabilities in content understanding and generation across diverse tasks~\cite{brown2020language,ouyang2022training}, such as question answering, summarization, and image generation. Their ability to handle multimodal inputs and generate outputs via prompts makes them a promising candidate for parameter synthesis. Furthermore, advancements in parameter-efficient fine-tuning and instruction tuning have enabled LLMs to adapt to downstream tasks more effectively.

Inspired by the work above, we are trying to create this new direction to utilize LLMs to generate nerual network parameters given data and instructions. This direction can also encourage potential research works to solve the real-world challenges, e.g., instrution-guided customized neural network acquisition and the scenario with limited training data.

To make preliminary exploration, we propose \texttt{NeuronGen}—a novel and easy-to-implement framework for neural network parameter generation using LLMs.  However, enabling LLMs to generate neural network parameters introduces several new challenges:
(1) There are no existing LLMs pretrained with an understanding of neural network parameters, nor existing methodologies for using such parameters as inputs for fine-tuning;
(2) It is non-trivial to prompt LLMs to generate nerual network parameters conditioned on data, task context and network architecture;

To tackle these issues, our approach adopts a two-stage training strategy:
stage 1: Parameter Referecne Knowledge injection and stage 2: Context-enhance Instruction Tuning. In Stage 1, we introduce neural network checkpoints and corresponding general instructions into pretrained LLMs to inject foundational knowledge of parameter structures. In Stage 2, we perform instruction tuning with task-specific data and enriched prompts that include information about the task, data, and neural networks. This context-aware training allows the model to learn flexible and adaptive generation strategies. Our main contributions are summarized as follows:
\begin{itemize}[leftmargin=15pt]
    \item To the best of our knowledge, this is the first work to explore the feasibility of directly generating neural network parameters using LLMs, without gradient-based optimization.
    \item We propose \ours, a novel, easy-to-implement, and extensible framework. Using a two-stage strategy, i.e., Parameter Reference Knowledge Injection and  Context-enhanced Instruction Tuning, we demonstrate that LLMs can generate functional neural network parameters.
    \item Our study introduces a new perspective: neural network parameters may exhibit latent structures tied to training context, which can be learned and reproduced by LLMs. This opens a new direction for understanding and interpreting neural network parameters through the lens of language models.
\end{itemize}
\section{Related Work}

\subsection{Neural Network Parameter Acquisition}

Typical model training paradigm based on backpropagation is commonly used approach and has demonstrated its success across different domains. Besides it, there are several other approaches about the nerual network acquisition. Ha et al.\cite{ha2016hypernetworks} propose using a hypernetwork—a smaller network that takes layer embedding vectors as input—to generate the parameters of a larger target network, referred to as the main network.
Finn et al.\cite{finn2017model} focus on meta-learning, where the goal is to train a model’s initial parameters so that it can quickly adapt to new tasks with just a few gradient updates using a small amount of data. 

More recently, several studies have explored using diffusion models to synthesize neural network parameters. Peebles et al.~\cite{peebles2022learning} construct a dataset of neural network checkpoints from multiple training runs, where each checkpoint includes model parameters along with metadata such as test loss and error for supervised tasks. Given an initial parameter vector and a target metric (e.g., loss or accuracy), their model learns to predict a distribution over updated parameter vectors for a fixed network architecture that achieves the target.
Similarly, Wang et al. train diffusion models to learn the distribution of high-performing parameters~\cite{wang2024neural}. They first use an autoencoder to map model parameters to a latent space, then train a diffusion model to generate these latent representations from random noise. 
Jin et al.~\cite{jin2024conditional} improve upon this by introducing a conditional latent diffusion model that synthesizes high-performing parameters based on specific task conditions.

\subsection{Large Lauguage Model}
Large language models (LLMs) have demonstrated remarkable capabilities across a wide range of domains, including natural language processing~\cite{min2023recent}, computer vision~\cite{wang2023visionllm}, bioinformatics~\cite{yang2022large}, and robotics~\cite{zeng2023large}. They have been successfully applied to various tasks such as question answering~\cite{dominguez2024questioning}, reasoning~\cite{plaat2024reasoning}, and data generation~\cite{wang2024survey}. In particular, substantial research has been devoted to LLM-driven image generation~\cite{bie2024renaissance}, text generation~\cite{li2024pre}, and code generation~\cite{jiang2024survey}.

Multimodal LLMs combine the advanced capabilities of LLMs with additional modalities such as images, audio, and video, enabling them to process information across multiple perceptual channels and mimic human multisensory cognition~\cite{survey_mml}. 
They typically incorporate modality-specific encoders (e.g., Vision Transformers for images) to fuse diverse inputs. 
This multi-modality enriches input representation and enhances both understanding and generation~\cite{BLIP,Flamingo,LLAVA}, supporting applications such as autonomous driving (sensor-based navigation), virtual assistants (e.g., Siri, Alexa), and medical diagnostics (e.g., combining blood tests with patient history for diabetes prediction).

\subsection{Key Differences from Existing Work}

Current approaches for neural network parameter generation typically focus on generating parameter distributions or initializations and often overlook how parameters can be adaptively generated in response to input data and tasks. Besides, few studies have explored the ability of LLMs to generate neural network parameters.  In contrast, our setting is fundamentally different: we leverage LLMs to generate model parameters in a prompt-driven, context-aware manner. Since our goal is to enable flexible, instruction-conditioned weight generation rather than optimize over fixed tasks or datasets, existing baselines are not directly applicable or comparable.




\begin{figure*}[t!]
\centering
    \includegraphics[width=1\textwidth]{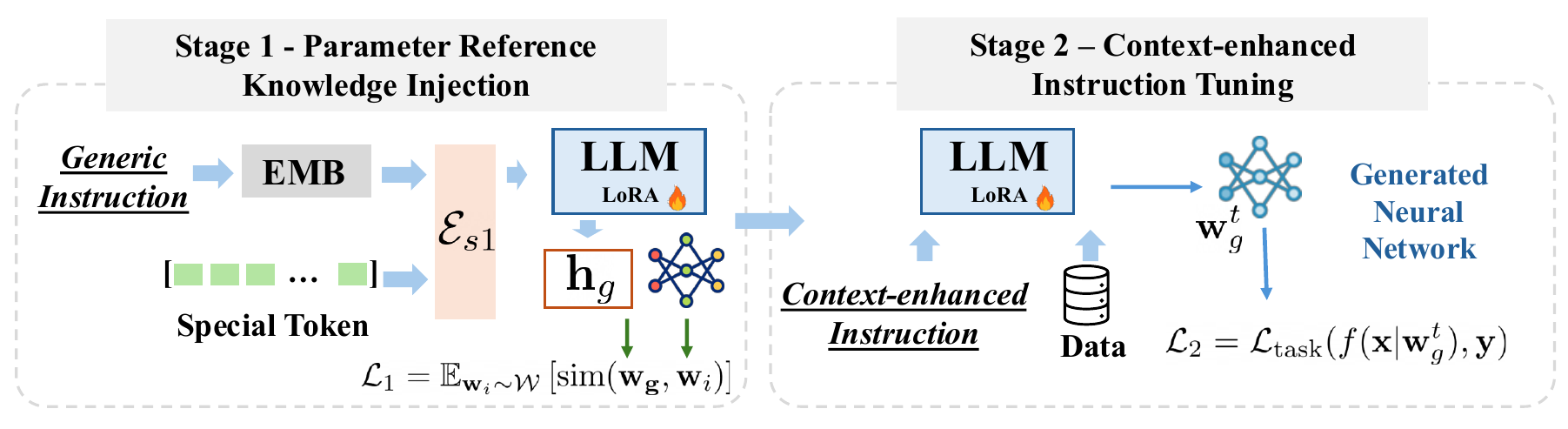}
    \caption{Framework of \ours. }
    \label{fig:framework}
\end{figure*}
\section{Methodology} \label{sec:method}
\subsection{Overview}
We design an approach for generating neural network parameters using LLMs with training context and prompted instructions, as illustrated in Figure~\ref{fig:framework} and Algorithm \ref{alg:method}. The framework consists of two stages: \textit{parameter reference knowledge injection} and \textit{context-enhanced instruction tuning}. In stage one, we aim to let the LLMs obtain a basic understanding of neural network parameters. To achieve this, we feed the neural network checkpoints of the target models into the LLMs denoted as $\mathcal{F}$ and conduct the knowledge injection via LoRA finetuning. In stage two, we aim to let the LLMs harness a deeper understanding of the target neural network and follow the text instructions of $\mathcal{F}$.
Thus, we send the training data, e.g., image data or text data, and more specific instruction into the LLM to conduct the context-enhanced instruction tuning. We describe each of these stages in detail in the following subsections.

\subsection{Notations}
Let $\mathcal{F}$ denote the LLM or Vision LLM. 
$\mathcal{P}$ represents the learnable special tokens, $\phi$ denotes the LoRA parameters, and $\theta$ refers to the projection layer parameters. 
Let \(f\) denote the architecture of the target model, whose parameter weights are to be generated by the LLM.  
We denote \(\mathbf{w}_i \sim \mathcal{W},\ i = 1, 2, \ldots\) as parameter weights of the model \(f\), sampled from a reference distribution \(\mathcal{W}\) (e.g., from conventionally trained checkpoints).  
\(\mathbf{w}_g\) represents the parameter weights generated by the proposed method for the same model architecture \(f\).
$\mathds{I}$ denotes the instruction input provided to the LLM. 
$\mathcal{D}$ represents the dataset corresponding to a specific task $t$, where each sample is a pair $(x, y)$ consisting of an input and its associated output.
Our task is to generate the parameters $\mathbf{w}_g$ of the target model \(f\), so that it fits the given task $t$ and dataset pairs $(x, y) \sim \mathcal{D}$.

\subsection{Stage 1: Parameter Reference Knowledge Injection} 
Similar to prior work~\cite{peebles2022learning,wang2024neural,jin2024conditional}, we aim to enable the LLM to understand the distribution of neural network parameters and generate parameters that align with it. However, this capability lies outside the scope of typical LLM pre-training data. To address this, Stage 1 focuses on injecting the missing parameter reference knowledge into the LLM.

\subsubsection{Input Preparation}


\noindent\textbf{Neural Network Parameter Preparation.} 
To capture and inject the distribution of neural network parameters \(\mathcal{W}\) into the LLM, we construct a dataset of neural network checkpoints obtained through standard gradient descent training. 
Specifically, we denote the collection of trained models as \(\{\mathbf{w}_1, \mathbf{w}_2, \ldots, \mathbf{w}_N\}\), where each \(\mathbf{w}_i\) represents a set of model parameters with same model structure obtained by training on the full dataset \(\mathcal{D}\) with a distinct random seed.

\noindent\textbf{Generic Instruction Construction.}
To better guide the LLM in developing a basic understanding of parameter reference knowledge, we provide generic instructions that do not require any specific information about the neural networks, training data, or the task. An example of such an instruction is \textit{``Please help generate parameters of neural networks.''}, denoted as \(\mathds{I}_{s1}\).


\noindent\textbf{Special Token Creation.}  
To enable parameter generation in parallel using a non-autoregressive approach, we feed special tokens into the LLM, together with the instruction, denoted as $\mathcal{P} \in \mathds{R}^{d_1 \times d_2}$. The dimensions $d_1$ and $d_2$ are chosen to satisfy the following conditions:  
(1) $d_2 = d_{\mathcal{F}}$, where $d_{\mathcal{F}}$ is the hidden dimension of the LLM;  
(2) $d_1 \times d_2 \ge |\mathbf{w}|$, where $|\mathbf{w}|$ denotes the total number of trainable parameters in the target neural network $f$ to be generated.

\subsubsection{Neural Network Alignment Learning}


\noindent\textbf{Primitive Parameter Generation.}
We feed the tuple consisting of the generic instruction \(\mathds{I}_{s1}\) and the learnable special token \(\mathcal{P}\), denoted as \(\mathbf{S}_1 = (\mathcal{P}, \mathds{I}_{s1})\), into the LLM \(\mathcal{F}\) for parameter generation.  
Note that the special token $\mathcal{P}$ is learnable; it is concatenated with the embedding of the first-stage instruction, \(\mathbf{e}_{s1} = \text{EMB}_{\mathcal{F}}(\mathds{I}_{s1})\), where \(\text{EMB}_{\mathcal{F}}(\cdot)\) denotes the embedding layer of \(\mathcal{F}\).  
The final input to the LLM for alignment learning is the concatenated representation \(\mathcal{E}_{s1} = [\mathcal{P}, \mathbf{e}_{s1}]\).

We then feed the representation \(\mathcal{E}_{s1}\) into the decoder of the LLM with LoRA applied, which is parameterized by \(\phi\). 
Let $\mathcal{H}_{s1} = [\mathbf{h}_g, \mathbf{h}_{s1}]$ denote the output hidden states, where $\mathbf{h}_g \in \mathbb{R}^{d_1 \times d_2}$ corresponds to the portion used for generating model parameters, and $\mathbf{h}_{s1}$ corresponds to the hidden states related to the input instruction. 
To match the dimensionality of the target neural network parameters \(|\mathbf{w}|\), we apply a projection \(\text{MLP}_{\theta}(\cdot): \mathbb{R}^{d_1 \times d_2} \rightarrow \mathbb{R}^{|\mathbf{w}|}\), parameterized by \(\theta\), to map the hidden representation to a flat vector of primitive model parameters $\mathbf{p}_{\text{pri}} = \text{MLP}_{\theta}(\mathbf{h}_g)$.
Finally, we apply a slicing operation to \(\mathbf{p}_{\text{pri}}\) to extract layer-wise parameters according to the architecture of the target model \(f\), resulting in the generated parameters \(\mathbf{w}_g\).

\noindent\textbf{Supervised Alignment Learning.}  
Following prior work~\cite{jin2024conditional,wang2024neural}, we inject knowledge of the target model parameter distribution into the LLM through supervised alignment learning.  
Our goal is to guide the LLM to generate model parameters \(\mathbf{w}_g\) that align with the reference distribution \(\mathcal{W}\).  
To this end, we sample \(N\) sets of model parameters \(\{\mathbf{w}_i\}_{i=1}^{N}\) from \(\mathcal{W}\), obtained by training the target neural network using a standard gradient-based optimization procedure.  
The generated parameters \(\mathbf{w}_g\) are then aligned with this reference set during training.
The alignment objective $\mathcal{L}_1$ is defined as:

\begin{equation}\label{eq:l1}
\mathcal{L}_1 (\mathcal{P},\phi,\theta) = \mathbb{E}_{\mathbf{w}_i \sim \mathcal{W}} \left[ \text{sim}(\mathbf{w}_i,
\mathbf{w_g}) \right], \quad  \mathbf{w_g} = \mathcal{F}(\mathcal{P},\mathds{I}_{s1}\ |\ \phi,\theta)
\end{equation}

where $\mathcal{F}(\mathcal{P}, \mathds{I}_{s1}|\phi,\theta))$ is the generated parameters $\mathbf{w}_g$, $\mathds{I}_{s1}$ is the instruction, $\mathcal{P}$ is the special token, $\phi$ is LoRA parameters, $\theta$ is the projection layer parameters, $\mathbf{w}_i$ represents reference parameters sampled from the distribution $\mathcal{W}$, and $\text{sim}(\cdot,\cdot)$ measures similarity between generated and reference parameters, such as negative mean squared error or cosine similarity.



\subsection{Stage 2: Context-enhanced Instruction Tuning} 

After Stage 1 parameter reference knowledge injection, the LLM has been exposed to the basic distribution of neural network parameters. However, it still lacks the ability to understand the underlying relationships between these parameters and their training context, such as training samples and the task descriptions. To address this, we introduce a second tuning stage aimed at improving the LLMs’ understanding of neural network parameters given the training data and detailed instructions. In the following subsections, we introduce how we create the input and conduct context understanding training for Stage 2.

\subsubsection{Triplet Input Construction}
For a given task $t$, we provide additional task-specific training data $\mathcal{D}^{t}$ and context-enhanced instructions $\mathds{I}^{t}_{s2}$. We construct a triplet input set $\mathbf{S}^{t}_2$ consisting of three components: a small, randomly sampled subset of training data $\mathcal{D}_{\text{sub}}^{t} \subseteq \mathcal{D}^{t}$, a task-specific instruction $\mathds{I}^{t}_{s2}$, and the special token $\mathcal{P}$, denoted as $\mathbf{S}^{t}_2 = (\mathcal{P}, \mathds{I}^{t}_{s2}, \mathcal{D}_{\text{sub}}^{t})$. The special token $\mathcal{P}$ remains the same as defined in Stage 1.

Compared with the generic instruction used previously, the instruction $\mathds{I}^{t}_{s2}$ provides a more detailed description of the task and dataset. Its format is:

\textit{“Please help generate parameters of the} \texttt{[\underline{Name of NN}]} \textit{neural network to conduct the \underline{classification} task}  \textit{with the} \texttt{[\underline{Name of Dataset}]} \textit{data samples.”}

An example of such an instruction is:

\textit{“Please help generate parameters of the} \texttt{[\underline{MLP}]} \textit{neural network to conduct the \underline{sentiment classification} task} \textit{with the} \texttt{[\underline{SST-2}]} \textit{data samples.”}



\subsubsection{Context Understanding Training Procedure}

\textbf{Context-Enhanced Parameter Generation.}
In this stage, we incorporate real data \(\mathcal{D}_{\text{sub}}^t\) and the context-enhanced instruction \(\mathds{I}^{t}_{s2}\) to further enhance the LLM’s ability to generate neural network parameters for a specific task \(t\).  
To achieve this, the triplet \((\mathcal{P}, \mathds{I}^{t}_{s2}, \mathcal{D}_{\text{sub}}^t)\) is fed into the LLM.  
Conditioned on the previously acquired understanding of parameter distributions, the model is guided to generate task-specific parameters \(\mathbf{w}^t_g\) that are well aligned with both the task \(t\) and the input data \(\mathcal{D}_{\text{sub}}^t\).

The processing of the instruction \(\mathds{I}^{t}_{s2}\) follows the same procedure as in Stage 1, where the learnable special token \(\mathcal{P}\) is concatenated in the embedding space.  
For textual input data, the processing is identical to that of the instruction.  
For other modalities, such as images, we employ the encoder associated with the multi-modal LLM to obtain their embeddings, ensuring that all inputs—including the instruction, data, and special token—reside in a shared representation space.


\textbf{Task-Specific Instruction Tuning.}
To guide the generation process to align with the specific task and data, we define a task-specific loss by evaluating the performance of the generated model \(\mathbf{w}_g^t\) on the given training data \(\mathcal{D}_{\text{sub}}^t\).  
The training objective of Stage 2 is formally defined as:

\begin{equation}\label{eq:l2}
    \mathcal{L}_2 (\mathcal{P},\phi,\theta) = \sum_{(\mathbf{x_j}, \mathbf{y_j}) \in \mathcal{D}_{\text{sub}}^{t}}\mathcal{L}_{\text{task}}(f(\mathbf{x}_j | \mathbf{w}^t_g), \mathbf{y}_j), \quad
    \mathbf{w}^t_g = \mathcal{F}(\mathcal{P}, \mathds{I}_{s2}, \mathcal{D}_{\text{sub}}^{t}\ |\ \phi,\theta)
\end{equation}
where \(f(\cdot \mid \mathbf{w}_g^t)\) denotes the target model instantiated with the generated weights \(\mathbf{w}_g^t\), and \(\mathcal{L}_{\text{task}}\) is the task-specific loss function, such as cross-entropy loss for classification tasks.  
\((\mathbf{x}_j, \mathbf{y}_j)\) represents a data pair from the given dataset \(\mathcal{D}_{\text{sub}}^t\).
It is important to note that the parameters \(\mathbf{w}_g^t\) are generated by the LLM \(\mathcal{F}\) and are not directly trained during the process.  
Instead, we optimize the set of parameters \(\{\mathcal{P}, \phi, \theta\}\) to enhance the LLM’s ability to generate effective neural network parameters \(\mathbf{w}_g^t\) by minimizing the loss \(\mathcal{L}_2\).

\SetKwInput{Input}{Input}
\SetKwInput{Output}{Output}
\SetKwProg{One}{\textcolor{blue}{Stage 1}}{}{}
\SetKwProg{Two}{\textcolor{blue}{Stage 2}}{}{}

\begin{algorithm}[ht]
    \SetAlgoLined
    \DontPrintSemicolon
    \Input{
    Data pairs $(x, y)$ selected from dataset $\mathcal{D}^t$ for task $t$ (e.g., classification).
    }
    \Output{
    Parameter $\mathbf{w}_g^t$ of model $f$.
    }
    \One{}{
    \For{$i=1,\ldots,N$}{
    Obtain reference parameter $\mathbf{w}_i$ based on the given dataset $\mathcal{D}^t$ via standard gradient-based training.
    }
    Construct LLM instructions $\mathds{I}_{s1}$ following Sec. 3.3.1\;
    \For{Training epoch $e=1,\ldots,E$}{
    Generate model parameter: $\mathbf{w_g} = \mathcal{F}(\mathcal{P},\mathds{I}_{s1}\ |\ \phi,\theta)$ \;
    $\mathcal{L}_1 (\mathcal{P},\phi,\theta) = \frac{1}{N}\sum_{i=1}^N \mathcal{L}_{mse} (\mathbf{w}_i, \mathbf{w}_g)$ \; 
    Update $\{\mathcal{P},\phi,\theta\}$ by minimizing loss $\mathcal{L}_1$ \;
    }
    }
    \Two{}{
    Construct LLM instructions $\mathds{I}_{s2}$ following Sec. 3.4.1 \;
    \For{Training epoch $e=1,\ldots,E$}{
    Generate model parameter that fits the given subset of data $\mathcal{D}^t_{\text{sub}}$: $\mathbf{w^t_g} = \mathcal{F}(\mathcal{P},\mathds{I}_{s2}, \mathcal{D}^t_{\text{sub}} \ |\ \phi,\theta)$ \;
    $\mathcal{L}_2 (\mathcal{P},\phi,\theta) = \sum_{(\mathbf{x_j}, \mathbf{y_j}) \in \mathcal{D}_{\text{sub}}^{t}}\mathcal{L}_{\text{CE}}(f(\mathbf{x}_j | \mathbf{w}^t_g), \mathbf{y}_j)$  \;
    Update $\{\mathcal{P},\phi,\theta\}$ by minimizing loss $\mathcal{L}_2$ \;
    }
    }
\caption{\text{Algorithm flow of \ours demonstrated with classification task.}}
\label{alg:method}
\end{algorithm}


\section{Experiments}



\subsection{Experiment Setups}
\noindent\textbf{Dataset and Task.} 
We assess the effectiveness of our proposed \ours using standard image and text classification tasks.  
For image classification, we use the benchmark datasets MNIST\cite{MNIST}, SVHN\cite{svhn}, and CIFAR-10\cite{cifar10}, all of which are 10-class classification tasks.  
For text classification, we use SST-2\cite{sst2}, SNLI\cite{snli}, and AG News\cite{ag-news}, which are binary, 3-class, and 4-class classification tasks, respectively.

\noindent\textbf{Implementation Details.} All experiments are conducted on an NVIDIA A100 with CUDA version 12.0, running on
a Ubuntu 20.04.6 LTS server. All baselines and the proposed FedType are implemented using PyTorch 2.0.1. We use Qwen2-VL-7B-Instruct\cite{wang2024qwen2} and Llama-3-8B-Instruct\cite{grattafiori2024llama} as our vision LLM and text-only LLM for the image and text tasks, respectively.
In Stage 1, we use mean squared error loss for parameter alignment.
In Stage 2, we use cross-entropy loss for classification tasks.
For the main experiments, we train for 30 epochs during the neural network alignment phase and 20 epochs for the context understanding phase. 
We follow the default train-test split provided by each dataset and use SGD optimizer for optimization.
The initial learning rate is set to \(10^{-3}\) and is halved every 10 epochs.


\noindent\textbf{Target Neural Network Model.} 
For image classification tasks, we consider the classic LeNet model\cite{Lenet} and a lightweight convolutional neural network (CNN) as target models. The lightweight CNN consists of three convolutional layers with ReLU activation and max pooling, followed by global average pooling and a fully connected layer.
For text classification tasks, we use lightweight Multi-Layer Perceptron (MLP) and Recurrent Neural Network (RNN) models. Both share a frozen embedding layer. The MLP performs mean pooling over token embeddings, followed by a ReLU-activated hidden layer and a final linear layer. The RNN is a single-layer vanilla RNN that uses the final hidden state for classification.
The details of the neural network structure can be found in the Appendix.


\subsection{Main Experiment Results}
In the main experiments, we follow the procedure described in Section~\ref{sec:method} to generate neural network parameters for both image and text classification tasks.  
The results are presented in Table~\ref{tab:main}.  
``Classical'' refers to the classification accuracy of models trained using standard gradient-based optimization, while \ours denotes models whose weights are generated by the LLM, using the same architecture as their classical counterparts.  
For a fair comparison, both the classical training and \ours use the full training set.

For image classification, we consider the well-known LeNet architecture and a lightweight CNN consisting of three convolutional layers with ReLU activation and max pooling.  
Under traditional training, both the lightweight CNN and LeNet perform well, achieving over 90\%, 80\%, and 60\% classification accuracy on MNIST, SVHN, and CIFAR-10, respectively. 
Although \ours does not fully match the performance of traditional training, it still achieves effective classification across all three datasets.  
On MNIST, \ours reaches accuracy comparable to the classical method.  
For more complex datasets like SVHN and CIFAR-10, performance drops slightly.  
\ours achieves over 60\% accuracy on SVHN with both the lightweight CNN and LeNet architectures.  
On CIFAR-10, it reaches 50\% accuracy with the CNN and 30\% with LeNet, indicating greater sensitivity to model capacity and dataset complexity.
For text classification, the proposed method achieves results comparable to the classical approach, except when generating the MLP classifier on the SNLI dataset.  
A possible explanation is that SNLI is not a straightforward classification task—it requires understanding the relationship between a premise and a hypothesis, which can be challenging for a simple MLP.  
Even the classical training method struggles to achieve high accuracy on this task.

Based on the main experimental results, we observe the following:  
(1) For simpler image datasets with lightweight models, the performance of neural networks generated by \ours can match or even exceed that of classical training—for example, on MNIST with both CNN and LeNet architectures.  
(2) For text classification tasks, \ours consistently achieves performance comparable to or better than the classical training approach.  
(3) When comparing image and text tasks, we find that overall performance on text datasets tends to be higher. This may be attributed to the strong text understanding capabilities of pre-trained LLMs.  
Overall, \ours demonstrates the promising potential of using LLMs for generating effective neural network parameters.




\begin{table*}[!h]
\caption{Experiment result comparison on image and text task.}
\resizebox{0.98\textwidth}{!}{
\begin{tabular}{l|c|c|c|c|c|c|c|c} 
\toprule 
\multirow{2}{*}{\textbf{Approach}}&\multirow{2}{*}{\textbf{NN}}&\multicolumn{3}{c|}{\textbf{Image}}&\multirow{2}{*}{\textbf{NN}}&\multicolumn{3}{c}{\textbf{Text}} \\
&  & \multicolumn{1}{c|}{\textbf{MNIST}} &
\multicolumn{1}{c|}{\textbf{SVHN}} & \multicolumn{1}{c|}{\textbf{CIFAR-10}} & & \multicolumn{1}{c|}{\textbf{SST-2}} &\multicolumn{1}{c|}{\textbf{SNLI}} & \multicolumn{1}{c}{\textbf{AG-NEWS}} \\
\midrule
 Classical & \multirow{2}{*}{{\textbf{CNN}}} &93.28&85.81&69.71&\multirow{2}{*}{{\textbf{MLP}}} & 74.31 & 44.59 & 88.42 \\
 \ours &  & 97.71 & 63.09 & 50.95 &  & 72.59 & 34.47 & 83.48 \\ 
 
 \midrule
 Classical &\multirow{2}{*}{{\textbf{LeNet}}}  & 99.01 & 89.27 & 62.05 &\multirow{2}{*}{{\textbf{RNN}}} & 77.63 & 61.74 & 84.72 \\
\ours & & 92.18 & 63.18 & 32.59 & & 76.03 & 59.27  & 85.14  \\
                                
\bottomrule 
\end{tabular}
}
\label{tab:main}
\end{table*}

\subsection{Ablation Study}

\begin{figure*}[h!]
\centering
    \includegraphics[width=1\textwidth]{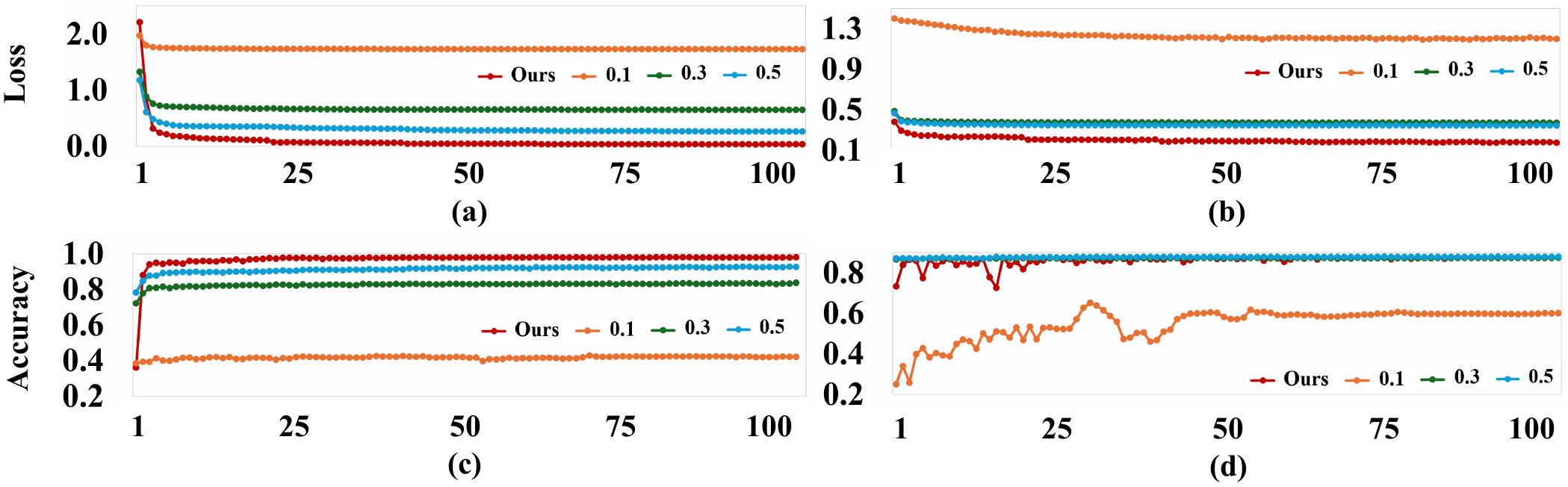}
    \caption{Ablation study on MNIST with CNN generation and AG News with MLP generation.  
    (a) and (b): Training loss of \ours versus Phase-2-only implementation on MNIST and AG News, respectively.  
    (c) and (d): Test accuracy of \ours versus Phase-2-only implementation on MNIST and AG News, respectively.}
    \label{fig:abl}
\end{figure*}

In this subsection, we examine the effectiveness of Phase 1, i.e., parameter reference knowledge injection. When we directly remove the Phase 1, we observe the generated parameters lose the basic bound and the logics outputed from the generated model is so massive that they cannot be optimized. In this case, we conduct a normalization to scale the parameters into different ranges controled by a hyperparameter $\alpha$. We demonstrate the results in Figure~\ref{fig:abl}.

We observe that the results with Phase 1 and Phase are able to reach the convergence faster and obtain a superior performance than the ones without Phase 1. One possible reason is that the Phase 1 provides some basic knowledge of the neural network parameters, thus it serves as a better initialization for the context-specific learning. Another observation is the different settings of soft clipping may affect the results in a certain range. But they all cannot overperform the full training process with Phase 1. On the other hand, we also find that our designed \ours with only Phase 2 still shows some very basic capbility. This indicates the feasibility and effectiveness of our designed training mechnism even without pre-knowledge injection.

\subsection{Model Generalization Exploration}

\begin{figure*}[t!]
\centering
    \includegraphics[width=1\textwidth]{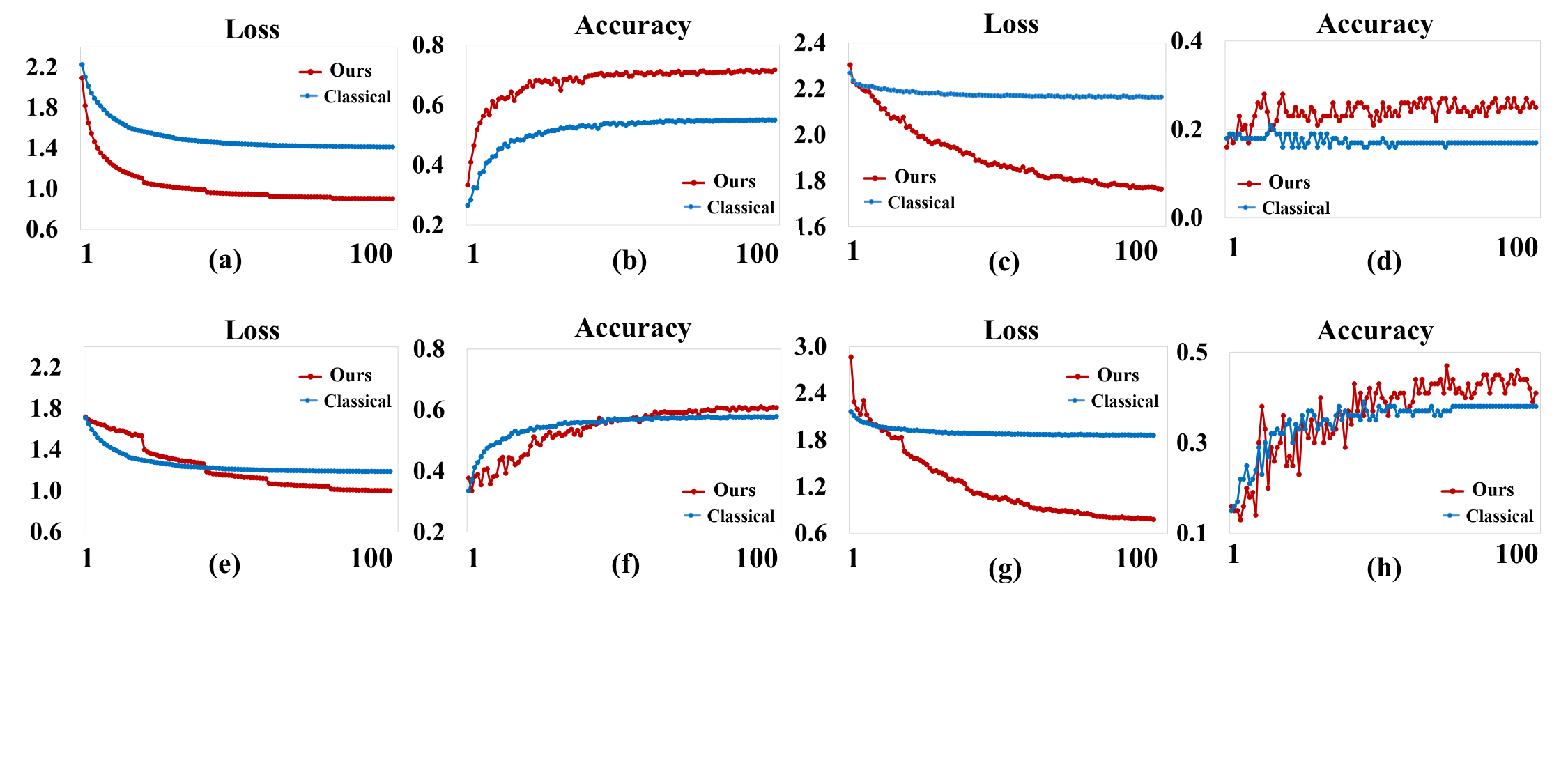}
    \caption{Model generation study on SVHN and CIFAR-10 datasets using CNN generation.  
    Training loss and accuracy on SVHN with (a–b) sufficient data and (c–d) limited data;  
    and on CIFAR-10 with (e–f) sufficient data and (g–h) limited data.
    }
    \label{fig:model_adp}
\end{figure*}

We further investigate the generalization capability of \ours.  
Given that \ours follows a two-stage learning process, it is capable of generating a target neural network \(\mathbf{w}_g^t\) for task \(t\).  
Here, we examine whether \ours can adapt to generate a new set of parameters \(\Tilde{\mathbf{w}}_g^t\) for the same task \(t\), but for a target model with a different architecture than \(\mathbf{w}_g^t\).
To test this, we generate parameters for a smaller CNN using an LLM that was pre-trained to generate parameters for a larger CNN.  
Specifically, we apply only Stage 2 (context-enhanced tuning) to generate the smaller model.  
In the experiment, the architecture of the smaller CNN is \texttt{[conv-ReLU-Pooling]} \(\times\) 2 \texttt{-MLP}, while the larger CNN used in pre-training follows the structure \texttt{[conv-ReLU-Pooling]} \(\times\) 3 \texttt{-MLP}.  
Figure~\ref{fig:model_adp} shows the loss and accuracy of the generated smaller CNN on SVHN and CIFAR-10, compared with the same model trained using the classical approach.

As shown in Figure~\ref{fig:model_adp} (a), (b), (e), and (f), when sufficient training data is available, \ours demonstrates strong performance on the SVHN dataset: it achieves steadily decreasing training loss and consistently higher test accuracy than the baseline across training epochs.  
For the CIFAR-10 dataset, the baseline initially reduces training loss more quickly and achieves higher accuracy. However, after around 50 epochs, \ours surpasses the baseline by achieving lower loss and ultimately higher test accuracy.
In the limited-data setting (10,000 training samples; see Figure~\ref{fig:model_adp} (c), (d), (g), and (h)), \ours outperforms the baseline on both SVHN and CIFAR-10.  
In these cases, \ours exhibits a sharper decline in training loss and yields better generalization, as reflected in higher test accuracy.


\subsection{Insights and Discussion}
Enabling LLMs to learn the ability to generate neural network parameters is an important yet largely unexplored area.  
In this work, we propose straightforward methods to investigate whether LLMs possess this capability and design a preliminary framework to support this goal.  
Based on our experimental results, we share the following insights with the research community to advance understanding in this emerging field.

\begin{itemize}[leftmargin=15pt]
    \item First, we demonstrate the feasibility of using LLMs to generate neural network parameters.  
    With appropriate guidance, the LLM is capable of generating model parameters that not only follow a target distribution but also adapt to specific tasks and real data—including both textual and visual modalities.  
    This is supported by our main experimental results in Table~\ref{tab:main}, where the generated models achieve strong performance on standard benchmarks and classic classification tasks.
    
    \item Second, we observe that LLMs have the potential to understand model structure, parameterization, and the process of learning parameters.  
    As shown in Figure~\ref{fig:model_adp}, an LLM trained on larger model architectures can generate effective smaller models, achieving lower loss and higher accuracy compared to models trained via standard optimization—especially in low-data regimes.
    This capability could benefit model compression and deployment applications, such as generating high-performing lightweight models for edge devices, where data is limited and classic training often underperforms.
\end{itemize}

\subsection{Limitation Discussion and Future Work}
While our proposed \ours demonstrates the feasibility of generating neural network parameters using LLMs, the current version still has several limitations.  
We outline these limitations below to highlight directions for future improvement:

\begin{itemize}[leftmargin=15pt]
    \item In our current approach, the LLM generates the entire set of neural network parameters in a non-autoregressive manner. That is, generating all model parameters at once.  
    This strategy does not scale well with increasing model size, as both the number of learnable special tokens and the size of the projection layer following LoRA grow proportionally with the target model.  
    When the target model is large and the available data is limited, the resulting optimization landscape becomes increasingly difficult to navigate.  
    As future work, we plan to incorporate parameter-efficient fine-tuning techniques—such as generating only the learnable prompts and LoRA parameters—for large and foundation models.

    \item In the current version of this work, we focus solely on classification tasks, as they provide an effective and straightforward way to assess the quality of parameter generation.  
    In future work, we plan to extend our investigation to generative tasks, which present additional challenges and opportunities for evaluating the generalization capabilities of LLM-generated models.

    \item Our current implementation relies on static and text-based descriptions of data. Real-world applications often involve multimodal or structured inputs, such as graphs, time-series signals, or visual data.  
    Incorporating richer, modality-specific representations or embeddings as part of the instructions and context could enable the LLM to better improve generation fidelity.  
    Future work will explore integrating structured or multimodal context encoding into the nerual network generation process.

\end{itemize}

\section{Conclusion}
In this paper, we propose \ours and made a priliminary exploration of adapting LLM to neural netwrok generation. By formulating parameter acquisition as a generative task, \texttt{NeuroGen} demonstrates the feasibility of a new paradigm in neural network development. Our two-stage design include Parameter Reference Knowledge Injection followed by Context-Enhanced Instruction Tuning—equips LLMs with both foundational and contextual understanding of neural parameters. Experimental results show that the generated parameters are not only structurally coherent but also functionally effective, suggesting that LLMs can internalize and express the latent mappings between training context and model weights. This work opens up a promising research direction at the intersection of generative modeling and network design, and paves the way for instruction-guided, data-efficient, and interpretable model generation in the future.

\bibliographystyle{unsrt}
\bibliography{egbib}
\newpage
\end{document}